\pdfoutput=1

\documentclass[11pt]{article}

\usepackage[]{ACL2023}

\usepackage{times}
\usepackage{latexsym}

\usepackage[T1]{fontenc}

\usepackage[utf8]{inputenc}

\usepackage{microtype}

\usepackage{inconsolata}

\usepackage{amsmath}
\usepackage{xspace}
\usepackage{xcolor}
\usepackage{booktabs}
\usepackage{graphicx}
\usepackage[most]{tcolorbox}
\usepackage{multirow}
\usepackage{colortbl}
\usepackage[linesnumbered, ruled]{algorithm2e}

\usepackage{cellspace} 
\newcommand{\AlgName}{\textsc{FastTrack}\xspace}

\title{\includegraphics[height=13pt]{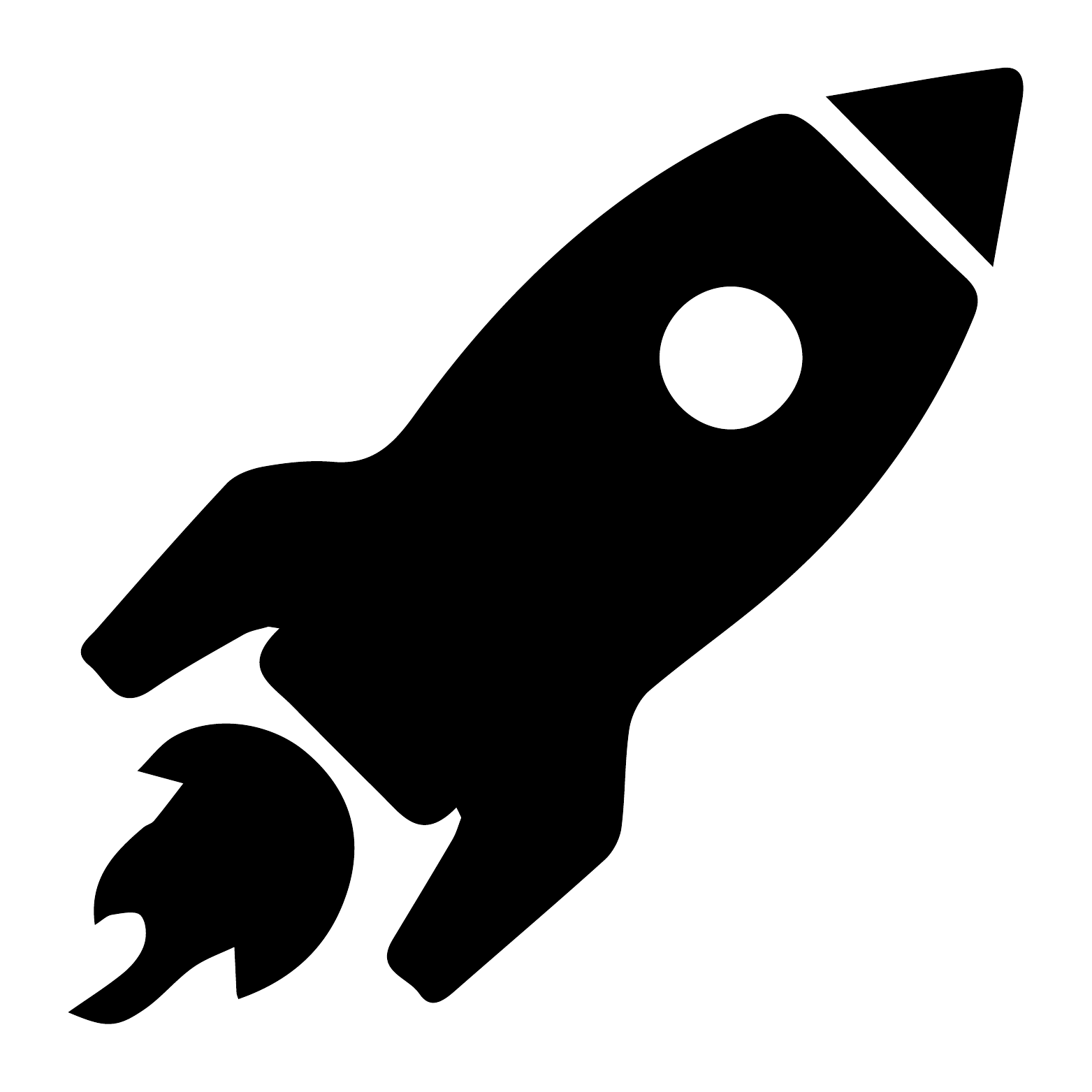} \AlgName: Fast and Accurate Fact Tracing for LLMs}

\author{Si Chen \\
  Virginia Tech \\
  \texttt{chensi@vt.edu} \\\And
  Feiyang Kang \\
  Virginia Tech \\
  \texttt{fyk@vt.edu} \\\And
  Ning Yu \\
  Netflix Eyeline Studios \\
  \texttt{ningyu.hust@gmail.com} \\\And
  Ruoxi Jia \\
  Virginia Tech \\
  \texttt{ruoxijia@vt.edu} 
  }

\begin{document}
\maketitle
\begin{abstract}

Fact tracing seeks to identify specific training examples that serve as the knowledge source for a given query. Existing approaches to fact tracing rely on assessing the similarity between each training sample and the query along a certain dimension, such as lexical similarity, gradient, or embedding space. However, these methods fall short of effectively distinguishing between samples that are merely relevant and those that actually provide supportive evidence for the information sought by the query. This limitation often results in suboptimal effectiveness. Moreover, these approaches necessitate the examination of the similarity of individual training points for each query, imposing significant computational demands and creating a substantial barrier for practical applications. This paper introduces \AlgName, a novel approach that harnesses the capabilities of Large Language Models (LLMs) to validate supportive evidence for queries and at the same time clusters the training database towards a reduced extent for LLMs to trace facts. Our experiments show that \AlgName substantially outperforms existing methods in both accuracy and efficiency, achieving more than 100\% improvement in F1 score over the state-of-the-art methods while being $\times33$ faster than \texttt{TracIn}.

\end{abstract}

\section{Introduction}

Recent years have witnessed \textit{large language models (LLMs)} demonstrating remarkable abilities in absorbing vast knowledge from extensive text corpora, yielding impressive advancements in NLP tasks such as question answering (QA). However, these models often produce seemingly coherent yet unfounded outputs, known as `hallucinations'~\citep{agrawal2023language}, posing risks in high-stake scenarios such as healthcare and finance, where reliability is of paramount importance \citep{MasterOfCode2023Hallucinations}. This critical challenge has motivated research on \emph{fact tracing}~\citep{akyurek2022towards}, which aims to identify the training data that serves as the knowledge source for LLMs' generation. Striving to provide a pathway to understanding and mitigating the issue of hallucination,
\citet{akyurek2022towards} proposed a benchmark for fact tracing, formulating it as a challenging task that involves searching for training data that has fact-support correspondence (i.e., supportiveness) with given queries. 
Current methods, however, tend to miss the mark and overly rely on similarity measures between individual training samples and the target query, such as gradient similarity~\citep{pruthi2020estimating, koh2017understanding}, embedding similarity~\citep{rajani2020explaining}, or lexical similarity~\citep{robertson1995okapi, lv2011lower}. 
As a natural result, these approaches may fail to differentiate between samples that merely look similar and those that actually contain the supporting information sought by the query–even in considerably simple cases. This prominent issue limits their effectiveness in identifying supportive training examples, preventing them from being effective in broader use cases~\citep{akyurek2022towards}. Besides, some of these methods, such as \citet{pruthi2020estimating, koh2017understanding}, carry a significant computational overhead in analyzing a given query. Providing intellectual inspiration for research exploration, nonetheless, its computational demand can be unaffordable for most practical scenarios. Despite soaring interest in this emerging problem, current research still falls short of the critical need by a large margin.

\begin{tcolorbox}[sharp corners, colback=white, colframe=black, boxrule=0.5pt, 
    top=1mm, bottom=1mm, left=1mm, right=1mm]
We summarize the \textbf{\textcolor{teal}{desiderata}} for fact-tracing methods as the following:
\begin{itemize}
    \item[$\diamond$] \textbf{\textcolor{teal}{D-i. Effective and Accurate}}. For a target query, fact-tracing methods need to identify \textit{all} supporting facts in the training corpus and achieve both high precision and recall simultaneously. 
    \vspace{0.5em}
\end{itemize}
\end{tcolorbox}
\begin{tcolorbox}[sharp corners, colback=white, colframe=black, boxrule=0.5pt, 
    top=1mm, bottom=1mm, left=1mm, right=1mm]
    \begin{itemize}
    \vspace{-0.2em}
    \item[$\diamond$] \textbf{\textcolor{teal}{D-ii. Computationally Tractable}}. Fact-tracing methods 
    need to be scalable with both the number of queries and the number of training samples to be examined.\vspace{-0.5em}
    \item[$\diamond$] \textbf{\textcolor{teal}{D-iii. Practically Robust.}} Fact-tracing prioritizes general-purposed, principled methods that are plausible for deployment and transferable between use cases. 
    \vspace{0.5em}
\end{itemize}
\end{tcolorbox}
Current methods all miss one or more of these principles. Specifically, gradient-similarity-based methods~\citep{pruthi2020estimating, koh2017understanding} are notoriously computationally demanding (\textcolor{teal}{\textbf{D-ii}}). Also, gradients are considerably susceptible to noises, rendering their performance rather unstable even with extensive hyper-parameter tuning~\citep{akyurek2022towards,park2023trak} (\textcolor{teal}{\textbf{D-i, D-iii}}). Lexical-similarity-based methods~\citep{robertson1995okapi, lv2011lower} are typically faster, but relying on queries and samples with supporting facts being similarly phrased. This assumption is not necessarily true in realistic use cases (\textcolor{teal}{\textbf{D-iii}}). Table \ref{tab:bm25-paraphrase} shows that the performance for such methods may drop a large margin under slight rephrasing of the text (\textcolor{teal}{\textbf{D-i}}). Therefore, these methods are neither practical nor reliable (as illustrated in Sec. \ref{box:fail-emb}). 

\begin{figure}[ht]
    \centering
    \includegraphics[width=0.45\textwidth]{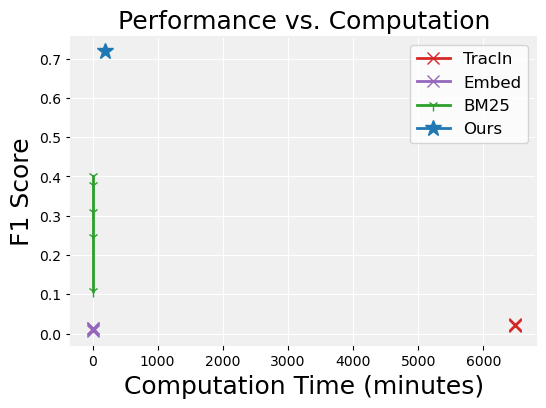}
    \caption{\emph{\AlgName achieves the best tradeoffs between fact tracing efficacy and efficiency.} The x-axis the the computational time of evaluating 100 queries using a 10k corpus, and the y-axis is the tracing performance when using top-k thresholds (if applicable). TDA methods yield consistently low performance across top-k thresholds, making them look like dots in the plot.}
    \label{fig:tradeoff}
\end{figure}

Determining whether a training example supports a factual statement in a query demands reasoning abilities beyond sample similarities where support for a factual assertion often arises through the inference of connections among related pieces of information. The dilemma with these approaches is that \textit{no single representation works in all cases} and the similarity in these pre-defined spaces may easily fail to capture the nuance of \textit{supportiveness} effectively. 
Inspired by the recent advancement in LLM's abilities in natural language understanding (NLU), a natural idea is to directly evaluate the supportiveness between each training sample and the target query using an LLM. Unprecedented in-context learning (ICL) capabilities make these models notably versatile and easily adaptable to novel cases with minimal customization, effectively bridging the realistic gap between fact-tracing methods and real-world scenarios. Admittedly, our preliminary investigation shows that this idea indeed enhances the efficacy in the identification of supportive training samples to an impressive extent. Nevertheless, this idea faces immediate challenges when applied to a practical-sized training corpora: traversal evaluation for all training sample-query pairs requires a massive number of queries to the LLM, unaffordable in both computation time and costs, hindering it from being practically useful. 

To address this dilemma, we propose \AlgName, which is a two-stage scheme decomposed into offline and online components. In the first stage, we build semantic indexes for the training corpus through hierachical clustering. Such process is completely offline and only need to be run once. During online stage, these pre-built semantic indexes facilitate the retrieval of relevant clusters for any given query, significantly reducing the search range. \AlgName then runs a fine-grained examination by employing a LLM to evaluate the supportiveness of training data in the retrieved clusters. While prior work \citep{akyurek2022towards} requires careful selection of small candidate set of size around 500 for practical evaluation, \AlgName enables a balance between computational feasibility and fine-grained analysis. This enables it to accommodate large corpus of size 10k or even 100k, while ensuring both satisfactory efficiency and efficacy (high precision and recall).

\textbf{Our contributions are summarized as follows:} 
\begin{itemize}\vspace{-0.5em}
    \item We propose a novel two-stage pipeline \AlgName and show it is easily adaptable without needing to train a model. (\textcolor{teal}{ \textbf{\textit{meets} D-iii}}) \vspace{-0.5em}
    
    \item We evaluate \AlgName's performance on various datasets with baseline methods. \AlgName achieves notable F1 scores of 0.72 on FTRACE-TREx and 0.91 on VITAMINC, more than \textbf{doubling} the performance of the best existing methods. (\textcolor{teal}{\textbf{\textit{meets}  D-i}})\vspace{-0.5em}
    
    \item We show \AlgName to offer a substantial edge in efficiency, being $\mathbf{33
    \times}$ \textbf{faster} than the TDA method \textsc{TracIn} for a corpus of 10k samples, and readily applicable to larger datasets with more than \textbf{100k samples.} (\textcolor{teal}{\textbf{\textit{meets} D-ii}}) \vspace{-0.5em}
\end{itemize}

\begin{figure*}[ht]
    \centering
    \includegraphics[width=1\textwidth]{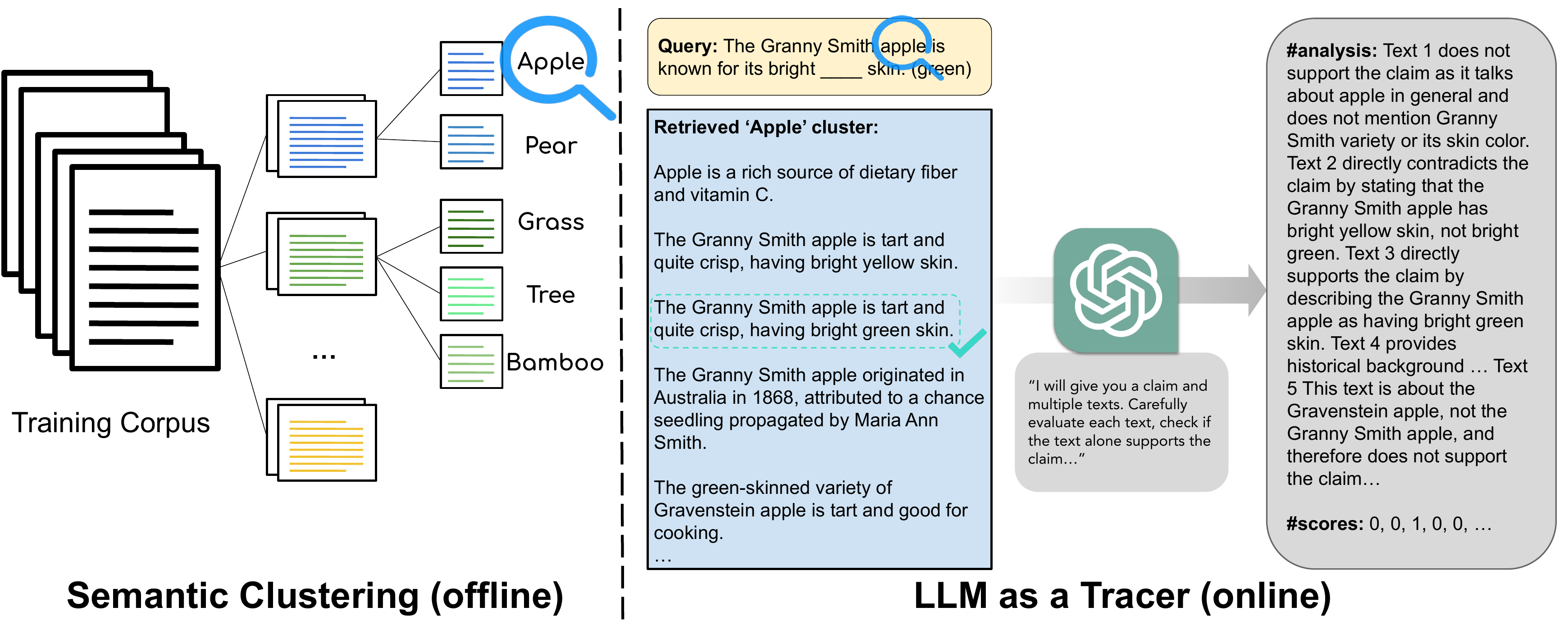}
    \caption{Illustration of \AlgName workflow. Stage 1, which is completely offline, reorganizes the training corpus into a semantic tree for easier navigation; Stage 2 retrieves relevant clusters using fuzzy keyword matching, then employs LLMs to assess candidate samples, retrieving those with a score of 1.}
    \label{fig:workflow}
\end{figure*}


\section{Related Work}
\paragraph{Training Data Attribution (TDA).}
TDA aims to trace model predictions back to the training examples that responsible for these predictions, which shares a similar goal with fact tracing. Prior work \citep{akyurek2022towards} proposes to use two main types of TDA methods as baselines: gradient-based and embedding-based attributions. 
Gradient-based methods, such as \textsc{TracIn} \cite{pruthi2020estimating}, estimate the attribution score of training data on predictions by calculating the cosine similarity between the gradients of the training data and the query. Embedding-based methods employs the model's internal representations to determine the relevance of training examples to a given test prediction \citep{rajani2019explain}. The attribution score is defined as a cosine product of hidden representations.

To retrieve supporting training data for a given query $z_\text{query}$, one need to score every training data and rank them by their influence score. As it could be computationally infeasible for gradient-based TDA scoring all training data in large datasets, \citet{akyurek2022towards} only evaluates on carelly selected small subsets (i.e., around 500) for each query. This limitation motivates us to design a framework that is both more computationally efficient and more effective.

\paragraph{Information Retrieval (IR).} 
IR focuses on retrieving relevant documents in a large collection given specific queries \citep{izacard2021unsupervised}. Though not originally designed for fact tracing task, prior work \citep{akyurek2022towards} found it effective and outperforms principled TDA methods by a large margin.
IR splits into two categories: term-frequency-based methods like BM25\citep{thakur2021beir,zhou-etal-2022-hyperlink}, which score each training data base on the token overlap with the given query, inversely weighted with the frequency of such tokens, and neural network-based methods \citep{izacard2021unsupervised,ni2021large}, which, despite their advanced capabilities, often require extensive manual annotations, making them less suited for fact tracing due to the absence of necessary annotations. Recent attempts to adapt neural methods through zero-shot learning have not matched BM25's performance \citep{thakur2021beir,zhou-etal-2022-hyperlink}. Therefore, following prior work, we select BM25 as the baseline for fact tracing due to its superior retrieval quality without the need for annotated data.


All of the methods above focus on \emph{relevance} while neglecting the \emph{supportiveness} of the connection between training data and the query. In this paper, we introduce \AlgName, the first supportiveness-aware approach for fact tracing, offering substantial benefits in real scenarios where training data may contain conflicting information.

\section{Methodology}
Fact tracing aims to identify knowledge source of a particular query. While similar to TDA, it focuses more on the fact-support correspondance between training data and query. This distinction is crucial: existing methods often retrieve relevant examples but fail to provide factual support, misaligning with the objective. The strong capability of LLMs such as ChatGPT makes it a perfect solution to provide justification based on `supportiveness'. However, directly doing pair-level comparison could be very time-consuming: Given a corpus of size $N$ and $m$ queries, the computation complexity is $\mathcal{O}(m N)$.

In this section, we introduce an original two-stage framework \AlgName, as illustrated in Figure \ref{fig:workflow}. In the first stage, \AlgName leverages a recursive clustering scheme to mine the semantic structure in the training corpus, which enables a coarse matching for a given query. This significantly refines the search range, making it feasible to perform a fine-grained examination of each candidate training examples in the second stage.



\subsection{Semantic Clustering} \label{subsec: cluster} The goal of the first stage is to create semantically meaningful indexes in an offline setting. This one-time process allows for the efficient utilization of these indexes in subsequent online stages, eliminating the need for re-computation. In this paper, we propose to employ a simple hierarchical clustering process over training data embeddings to recover underlying tree structures of the data. This process reorganize the entire training corpus into a more structure format, laying the groundwork for more effective data navigation and retrieval.  We first apply k-means clustering on the sample embeddings to mine its semantic structure. The clustering is conducted recursively where larger clusters will be further clustered until the size of all the clusters is within a certain threshold.

The key of our method lies in transcending the limitations of conventional clustering algorithms, which typically do not assign semantically meaningful labels to each cluster. By harnessing the power of Large Language Models (LLMs), \AlgName assigns a carefully selected set of keywords to each cluster, serving as its semantic label. This strategic integration not only renders the clustering outcomes interpretable but also significantly simplifies the process of navigating through the corpus in response to specific queries. We note that such semantic clustering only need to be applied once offline, effectively allowing us to leverage the massive amount of compute in pre-training for free.

\subsection{LLM as a Sample-Level Tracer} \label{subsec: sample}
With the structured and semantically meaningful clusters, we can now online process each query for fact tracing efficiently. The first step is to retrieve relevant clusters for a given query. A simple example for such cluster-level retrieval is to apply fuzzy match \footnote{\url{https://github.com/seatgeek/thefuzz}} to identify those clusters that shared similar keywords as the query. Furthermore, the efficacy of clustering can be enhanced through ensemble of different clustering outcomes, as detailed in Table \ref{tab: ensemble}. 

Now, with the retrieved clusters, the second step is to identify the groundtruth supporting data from this narrowed pool. We frame this stage as a binary verification problem: given a specific query, we classify each candidate training example into two categories based on its `supportiveness'. An example is considered 'grounding' if it supports the query. 
A direct way to perform such classification is to instruct the LLM to evaluate a single training example against a query for supportiveness, assigning a score of 1 for supportiveness and 0 otherwise. Although effective, this one-at-a-time scoring method can still be computationally and financially costly. To futher enhance efficiency and speed up the process, we devised the prompting strategy to evaluate a batch of training data in a single inference run. This batch processing approach significantly cuts down the time required for evaluations, reducing the number of necessary inferences by a factor of $b$, where $b$ is the number of candidate examples in a batch. The example prompt used in our experiments can be found in Appendix \ref{sec:prompt}.
Following the LLM's evaluation, examples that are assigned a score of 1, indicating supportiveness, are systematically retrieved. The detailed workflow of \AlgName is presented in Algorithm \ref{alg:workflow}.


\section{Experimental Setup}
\subsection{Datasets}

\paragraph{FTRACE-TREx.}
The FTRACE-TRex dataset is proposed by \citep{akyurek2022towards}, with 27k queries created using LAMA \citep{petroni-etal-2019-language} and 1M masked training examples extracted from TREx \citep{elsahar2018t} as the attribution set. Each training example is a cloze-style sentence with either the subject or object masked. The groundtruth training example for each query is defined as the examples that express the same fact, regardless of the masking position. To address the computational overhead, \citet{akyurek2022towards} proposes to construct a small, separate candidate set for each query (around 500). We follow a similar setup, but use a \textbf{larger, fixed} candidate pool to better reflect real-world scenarios: we randomly sample 100 queries from the entire query set for evaluation, and build the candidate pool by including all the corresponding groundtruth,   supplementing with random samples to form a corpus of \textbf{10k}. 

\paragraph{VITAMINC.}
We incorporate the VITAMINC dataset \citep{schuster-etal-2021-get} as a means to evaluate fact tracing methods' ability to mirror real scenarios where training corpus of LMs containing contradictions or misinformation. 
The VITAMINC dataset is built based on factual revisions to Wikipedia: each single factual revision yields a contrastive pair of contexts, where one context refutes the given claim and the other supports it. The original VITAMINC dataset presented each entry in the format of \emph{claim}, \emph{evidence}, and \emph{label}, where the label indicates if the evidence 'SUPPORTS', 'REFUTES', or provide 'NOT ENOUGH INFO' to the evidence. To use it for fact tracing purposes, we build the attribution set by collecting 10k unique pieces of evidence (acting as training data). Then the query set is built by collecting corresponding claims that can be supported by the evidence.
\footnote{Due to the labeling format of the original dataset, some claims may have more than one supporting evidence but we do not know. To address such an issue, we manually inspect 100 queries for their groundtruth data and use these queries for evaluation. We provide the data we manually inspect along with this submission.}

\subsection{Baselines}
Following \citet{akyurek2022towards}, we compare our method \AlgName with TDA methods (i.e., \textsc{TracIn}, \textsc{Embed}) and the most representative IR method (i.e., BM25).
\paragraph{\textsc{TracIn}.}
\textsc{TracIn} \citep{pruthi2020estimating} is a recent gradient-based TDA method that has demonstrated strong empirical results and tractability. Following the setup of \citet{akyurek2022towards}, we use an optimized version of \textsc{TracIn} by rescaling gradients with Adafactors accumulators, applying unit-normalization to the gradients, and selecting the best-performing layer. Data in FTRACE-TREx are cloze-style examples, hence we finetune an MT5 model \citep{xue-etal-2021-mt5} following \citet{akyurek2022towards} to predict the masked tokens. We note that gradient similarity is only meaningful when query and training data have the same question-answer construction, and it is difficult to construct the VITAMINC dataset in this way. Hence, we omit the evaluation of \textsc{TracIn} on VITAMINC dataset.

\paragraph{\textsc{Embed}.}
Embedding-based similarity is another popular branch for fact tracing tasks. Here we refer to Equation \ref{eq: embed} as baseline \textsc{Embed}. For FTRACE-TREx dataset, we use the same fine-tuned MT5 model as for \textsc{TracIn}, selecting the best-performing layer as the final result. For the VITAMINC dataset, we finetune a BERT model \citep{kenton2019bert} on our constructed attribution set.

\paragraph{BM25.}
We use a publicly available implementation of BM25 \citep{lv2011lower} as our baselines \footnote{\url{https://pypi.org/project/rank-bm25/}}. We tokenize queries and training examples by space, removing any masked tokens. We proceed with the default settings for all hyperparameters, ensuring a standardized approach for our baseline comparisons.

\subsection{Tracing Performance Evaluation}
TDA methods and BM25 score a given test query against every training example and then sort all examples based on their scores. This results in a top-k precision and recall performance measurement, where the k is the threshold of taking the top k ranked examples as the retrieved supporting training data \citep{akyurek2022towards}.
In contrast, our method directly retrieves the supporting training data without ranking. To enable a unified comparison, we use F1 score as the main metric. We report the best-performing F1 score and the corresponding precision and recall for each method. 

\section{Empirical Results}

\subsection{Overall Performance}
We first evaluate the overall performance of different methods on FTRACE-TREx and VITAMINC datasets in Table \ref{tab:res-main}. Hyperparameters for all methods are presented in Appendix \ref{sec:baseline-details}.

\begin{table}[h]
    \centering
    \caption{Comparison of fact tracing performance. We present the best F1 scores among top-k for each method; precisions and recalls are chosen at the threshold lead to optimal F1 score. Among all methods, \AlgName performs the best. *The last row gives the upper bound performance achievable in the first cluster-level retrieval stage. }
    \label{tab:res-main}
    \scalebox{0.7}{
    \begin{tabular}{Sc Sc Sc Sc Sc Sc Sc}
    \toprule
                    & \multicolumn{3}{c}{\textbf{FTRACE-TREx}}          & \multicolumn{3}{c}{\textbf{VITAMINC}} \\ 
    \cmidrule(lr){2-4} \cmidrule(lr){5-7}
                    & \textbf{F1} & \textbf{Precision} & \textbf{Recall} & \textbf{F1} & \textbf{Precision} & \textbf{Recall} \\
    \midrule
    \textbf{\textsc{TracIn}} & 0.02        & 0.19               & 0.01            & -           & -                  & -               \\
    \textbf{\textsc{Embed}}  & 0.01        & 0.08               & 0.01            & 0.48        & 0.54               & 0.46            \\
    \textbf{BM25}   & 0.40        & 0.49               & 0.52            & 0.55        & 0.59               & 0.53            \\
    \textbf{Ours}   & \textbf{0.72}        &  \textbf{0.81}               &  \textbf{0.69}            &  \textbf{0.91}        &  \textbf{0.88}               &  \textbf{0.98}            \\
    \textbf{Ours\textsuperscript{*}} & 0.86        & 0.92               & 0.83            & 1.00       & 1.00               & 1.00            \\
    \bottomrule
    \end{tabular}}\vspace{-1.5em}
\end{table}

Fact tracing is a challenging task. Previous work \citep{akyurek2022towards} proposes several techniques to optimize TDA methods but found that even BM25 with no tuning outperforms TDA, and all these methods are far from perfect. In Table \ref{tab:res-main} we show similar findings, where \textsc{TracIn} and \textsc{Embed} resulted in F1 score lower than $0.1$ on FTRACE-TREx dataset. We also observe that \textsc{TracIn}'s performance is highly dependent on the chosen model checkpoint. Specifically, the performance noted in our main results table was achieved using the final 80k-step checkpoint, with earlier checkpoints yielding even weaker outcomes (as shown in Appendix \ref{sec:res-more}).

\begin{tcolorbox}[sharp corners, colback=white, colframe=black, boxrule=0.5pt, 
    top=2mm, bottom=2mm, left=2mm, right=2mm]
\textbf{\textcolor{teal}{Takeaway:}} 
\AlgName delivers impressive tracing performance, yielding both high precision and recall, improving the F1 score by \textbf{>80\%} compared to the best-performing baseline BM25.

\end{tcolorbox}

All baseline methods retrieve training examples based on their `relevance' to the given query, which could violate the goal of fact tracing. This discrepancy becomes evident in real-world scenarios, where datasets, unlike the scientifically accurate and consistent ones often evaluated in prior research, contain conflicting information. Our evaluation on VITAMINC dataset reveals that such methods yield low precision due to their relevance-focused logic. Notably, \AlgName significantly outperforms all baselines, achieving an F1-score of 0.91, demonstrating its effectiveness in accurately identifying grounding training data for queries.

\begin{tcolorbox}[sharp corners, colback=white, colframe=black, boxrule=0.5pt, 
    top=2mm, bottom=2mm, left=2mm, right=2mm]
\textbf{\textcolor{teal}{Takeaway:}} \AlgName not only excels in fact-tracing performance but also offers the optimal balance between computational speed and effectiveness. It outperforms competitors significantly, running \textbf{33 times faster} than \textsc{TracIn} in evaluating 100 queries (Figure \ref{fig:tradeoff}).
\end{tcolorbox}

\subsection{Failure Analysis}\label{subsec:qualitative} 
In this section, we qualitatively examine some failure examples of different tracing methods to shed light on the future direction of fact tracing.

\paragraph{When does BM25 fail?}
BM25 operates based on token overlap, and retrieves examples with high lexical similarity to the query, regardless of their factual consistency. As shown in the example below, while the first retrieved example is correct, the second contradicts the query, and the third is entirely unrelated.
\begin{tcolorbox}[rounded corners, colback=lightgray!20, colframe=black, boxrule=0.5pt, 
    top=2mm, bottom=2mm, left=2mm, right=2mm]
\small
\textbf{Query}: Alloy Digital's network has a monthly reach of more than 100 million unique visitors. \\[2ex]
\textbf{BM25 Retrieved:}\\
\textbf{Rank-1}: Defy Media: According to comScore, Alloy Digital's network reaches over 221 million unique visitors each month, including more than half of the aged 12-34 internet users.\\
\textbf{Rank-2}: According to comScore, Alloy media platforms reach over 95 million unique visitors each month, including over half of the age 12-34 internet users.\\
\textbf{Rank-3}: The franchise has sold more than 26 million units worldwide with the release of 2018 's installment.
\end{tcolorbox}

BM25's performance can be poor even when there are no such data conflicts. We further conduct experiment on FTRACE-TREx dataset where we paraphrase each query using an open-sourced paraphraser \footnote{\url{https://huggingface.co/humarin/chatgpt_paraphraser_on_T5_base}}. The performance of BM25 before and after paraphrasing is shown in Table \ref{tab:bm25-paraphrase}, where both precision and recall drop by a wide margin.

\paragraph{When do TDA methods fail?}

\textsc{TracIn} conducts a first-order approximation and uses the dot product of the model's gradients between each train-test sample pair to measure this contribution
However, we find its actual performance is fragile and can be affected by a number of factors.

\emph{1) \textsc{TracIn}'s performance is highly dependent on having the exact same construct of question-answer pairs.}
LMs for QA tasks typically use an encoder-decoder architecture, such as T5/MT5. The gradient is calculated with respect to the loss of the word/token being predicted. However, gradient similarity between a train-test sample pair is only meaningful when these are the same QA questions with identical question-answer pairs. In other words, even for sample pairs where the texts are the same, if the construction of question-answer is different, the loss and gradient may be unrelated. 
This aligns with our evaluation results: we find that \textsc{TracIn} cannot identify those groundtruth training examples with supporting facts but having different QA construction. This results in arbitrarily poor performance on some queries, as the cosine similarity between gradients - which are high-dimensional vectors - can be dominated by unrelated factors and fail to capture the actual correlation between samples.

\emph{2) \textsc{TracIn} tends to retrieve sentence with the same masked token.} 
Such finding has also been observed in \citep{akyurek2022towards}. This likely occurs because the same masked token produces similar training gradients.

\begin{tcolorbox}[rounded corners, colback=lightgray!20, colframe=black, boxrule=0.5pt, 
    top=2mm, bottom=2mm, left=2mm, right=2mm]
\small
\textbf{Query}: Comptroller of Maryland is a legal term in \_\_\_\_. (Maryland) \\[2ex]
\textbf{\textsc{TracIn} Retrieved:}\\
\textbf{Rank-1}: The \_\_\_\_ Comptroller election of 2010, was held on November 2, 2010. (Maryland)\\
\textbf{Rank-2}: It is found in Alabama, Florida, Louisiana, \_\_\_\_, Mississippi, North Carolina and Virginia. (Maryland)\\
\end{tcolorbox}
\noindent As illustrated in the example above, the top-ranked retrieved example is correct, where the training example and query share the same masked target token. However, the second retrieved example does not provide any relevant fact, only the masked token to predict is the same.

The other TDA method evaluated in this paper, \textsc{Embed}, relies on hidden space similarity search. The dilemma for this approach is that \textit{no single representation works for all tasks}~\citep{vaze2023no}, which is more pronounced in these QA problems. The similarity of text pairs could be measured from different perspectives and the one that is best captured does not necessarily focus on the "supporting fact". Another major issue with this approach is that similar texts always receive similar scores, rendering the results end up in clumps. If the front-running clump is wrong, all samples in the clump are wrong, yields zero top-k accuracy. For example, for the same query "\emph{Comptroller of Maryland is a legal term in <MASK>}", the top 3 retrieved examples of \textsc{Embed} are:
\begin{tcolorbox}[rounded corners, colback=lightgray!20, colframe=black, boxrule=0.5pt, 
    top=2mm, bottom=2mm, left=2mm, right=2mm, label={box:fail-emb}]
\small
\textbf{Rank-1}: the Mayor of \_\_\_\_. (Moscow)\\
\textbf{Rank-2}: Embassy in Cyprus is located in \_\_\_\_. (Nicosia)\\
\textbf{Rank-3}: He served on the \_\_\_\_ of Edmonton. (town council)
\end{tcolorbox}
These retrieved examples, to varying degrees, relate to the query by involving 1) public offices and elected officials, 2) political or geographical entities, and 3) individuals with governmental roles. In fact, The groundtruth example belongs to a similar category. Yet, embedding similarity cannot detect fact-support correspondence between samples and cannot distinguish different levels of sample similarities.


\section{Ablation Study and Analysis}
\paragraph{In-depth Analysis of \AlgName.}
The first stage of \AlgName - cluster level retrieval - decides the performance upper bound of our methods. If relevant clusters are not identified during this phase, it becomes impossible to recover them in the later stage.  We report the upper bound performance achievable in the last row of Table \ref{tab:res-main}, to reveal the limitation origins from the first stage. Specifically, this upper bound assumes perfect accuracy in the second stage, meaning if the correct cluster is identified, we achieve 100\% precision and recall on this cluster. As shown in Table \ref{tab:res-main}, the upper bound of \AlgName has a short gap to perfect. The precision is 0.92 while the recall is only 0.83. Such failure origins from the first stage could come from two sides:

\emph{1) clustering algorithm.} The clustering algorithms group data with similar embedding together. Although in general, we observe that groundtruth training data for a specific query usually falls within 4 clusters on average, which means the clustering algorithm successfully groups relevant training data into the same cluster, there still exists the case that for some clusters the groundtruth training data is the minority. In such case, groundtruth data could be ignored when assigning the cluster semantically meaningful keywords, making this cluster hard to retrieve. In practice, this can be improved by using an ensemble - we observe that an ensemble of three yields a performance upper bound of precision 0.92, recall 0.83; while single clustering yields an upper bound of precision 0.81, recall 0.65.

\begin{table}[htbp]
\centering
\caption{Upperbound performance of \AlgName when using single and ensemble embeddings on FTRACE-TREx.}
\label{tab: ensemble}
\scalebox{0.8}{
\begin{tabular}{cccc}
\hline
\textbf{}          & \textbf{Single} & \textbf{Two-Ensemble} & \textbf{Three-Ensemble} \\ \hline
\textbf{Precision} & 0.81            & 0.89                  & 0.92                    \\
\textbf{Recall}    & 0.65            & 0.78                  & 0.83                    \\ \hline
\end{tabular}}
\end{table}

\emph{2) cluster retrieval method.} We currently employ simple fuzzy matches to capture clusters that share similar keywords as the query. However, the training data may present the query on a different surface. Future studies could leverage more advanced tools to enhance the process.

Table \ref{tab:res-main} shows that there exists a gap between performance upper bound and final performance. This gap comes from ChatGPT's limitation, where it misclassified a few examples. We show two interesting types of misclassification here:
\begin{tcolorbox}[rounded corners, colback=lightgray!20, colframe=black, boxrule=0.5pt, 
    top=2mm, bottom=2mm, left=2mm, right=2mm]
\small
\textbf{Query}: President of the Executive Yuan is a legal term in \_\_\_\_\_. (Taiwan)\\[2ex]
\textbf{False negative examples (mask removed)}:\\
1. He has interviewed financial services regulators including Sean Chen (politician), the Premier of Taiwan, when he was the Chairman of the Financial Supervisory Commission (Republic of China) of Taiwan and negotiated the financial Memorandum of Understanding with China.\\
2. Hsich Tung-min was the ninth Governor of Taiwan Province (1972-1978) and the sixth and first local Taiwanese Vice President of the Republic of China (1978-1984) under President Chiang Ching-Kuo.\\[2ex]
\textbf{GPT-4 analysis:}\\
The term "President of the Executive Yuan" is not mentioned in any of the texts. The texts mention various political positions in Taiwan, such as the Premier of the Republic of China and the President of Taiwan, but none of them refer to the President of the Executive Yuan. Therefore, it cannot be inferred from the texts that "President of the Executive Yuan" is a legal term in Taiwan.
\end{tcolorbox}
In the above example, GPT-4 did not recognize that the 'Executive Yuan's leader is the `Premier of Taiwan', indicating a gap in connecting related concepts. The second failure example appears to be a labeling error. Another example is that GPT-4 struggles with complex logical reasoning involving dates; for instance, it incorrectly equates the information from different dates, focusing merely on numerical comparisons (see Appendix \ref{sec:res-more}). 
Failure cases at this stage mainly stem from LLM's own bottleneck. These challenges represent a significant area of ongoing research and are beyond the scope of our current study. We acknowledge these limitations and suggest them as critical avenues for future investigation to enhance the capabilities and applications of LLMs.

\paragraph{Embeddings Schemes.}
We use Sentence-Transformer \footnote{\url{https://www.sbert.net}} as the embedding model to perform clustering in our main evaluation. To test the sensitivity of \AlgName on different choices of embeddings, we also test some state-of-the-art embedding models such as Cohere Embed v3\footnote{\url{https://txt.cohere.com/introducing-embed-v3/}} and Mistral-Embed\footnote{\url{https://docs.mistral.ai/api/}}.
As shown in Table \ref{tab: emb},  \AlgName consistently achieves comparable top-performance upperbounds across various embedding models, underscoring its adaptability to different embedding choices.



\paragraph{Corpus Size.}
Moving forward, we aim to tackle a more challenging scenario: we use the same query set of VITAMINC, but augment the attribution set with additional non-relevant examples until the total reaches 100k. This setting is designed to evaluate our method's robustness in scenarios that better resemble real-world applications. As shown in Table \ref{tab:corpus}, both methods exhibit a slight decline in performance, yet \AlgName consistently outperforms BM25 by a significant margin. BM25's performance drop is ascribed to the inclusion of new examples that exhibit high lexical overlap with the queries, while our method, mainly stems from the clustering stage, where the clustering logic has been impacted after a more diverse sample are included. We leave a detailed analysis in Appendix \ref{sec:res-more}.


\begin{table}[t]
\centering
\caption{Performance of BM25 and \AlgName when dealing with different corpus size. Both of the methods encounter a slight performance drop, while \AlgName is still $1.66\times$ better than BM25.}
\scalebox{0.7}{
\begin{tabular}{ccccccc}
\hline
\textbf{}      & \multicolumn{3}{c}{\textbf{VITAMINC-10k}}          & \multicolumn{3}{c}{\textbf{VITAMINC-100k}}         \\ \cmidrule(lr){2-4} \cmidrule(lr){5-7}
\textbf{}      & \textbf{F1} & \textbf{Precision} & \textbf{Recall} & \textbf{F1} & \textbf{Precision} & \textbf{Recall} \\ \hline
\textbf{BM25}  & 0.55        & 0.59               & 0.53            &  0.53           & 0.56                   & 0.50                \\
\textbf{Ours}  & 0.91        & 0.88               & 0.98            &  0.88           &   0.85                 & 0.92               \\
\textbf{Ours\textsuperscript{*}} & 1.00        & 1.00               & 1.00            &  0.95           & 0.95                   & 0.95                \\ \hline
\end{tabular}}
\label{tab:corpus}
\end{table}\vspace{-0.5em}

\section{Conclusion}
In this paper, we introduced \AlgName, a pioneering two-stage framework designed to address the shortcomings in current fact tracing methodologies, particularly their neglect of the 'supportiveness' of evidence. \AlgName substantially improves the tracing performance by more than 100\% in F1 score, and offers computational efficiency, capable of handling large datasets up to 100k in size. We also provide a thorough analysis of each tracing method to shed light on future direction in fact tracing. It's important to note that the current performance bottleneck primarily stems from the limitations of GPT models. Therefore, future efforts could focus on fine-tuning a model specifically for tracing purposes.





\section*{Limitations}


While our proposed method, \AlgName, has shown considerable success, it's important to acknowledge that its performance is ultimately constrained by the capabilities of GPT models. Thus, future work could explore techniques to fine-tune a LLM specifically targeting tracing purpose. Another limitation of \AlgName is its capacity to process only a limited number of training examples in each batch. This presents an opportunity for future improvements by incorporating techniques that can handle longer contexts. By doing so, it may be possible to decrease the necessity for multiple inferences, thereby optimizing the process.


\section*{Ethics Statement}

Our research advances the accuracy and efficiency of fact tracing techniques, elucidating the connections between the training data of LLMs and their generated assertions. Our method ensures data privacy integrity, as it solely utilizes publicly accessible data samples, and is not designed for the inadvertent generation of unintended content by LLMs. While mindful of the potential for redistributing web data to inadvertently disseminate misinformation, we foresee no other ethical concerns with our methods.

Committed to the ethos of open science, our study champions reproducibility, transparency, and the facilitation of further inquiry. To this end, we will grant unrestricted access to all materials related to our research. This includes a comprehensive, meticulously documented repository encompassing all scripts, models, and codes necessary for preprocessing and evaluation, allowing for the full replication of our experiments. Beyond making these resources available, we are dedicated to their ongoing maintenance and to providing timely support for any inquiries or clarifications. This commitment underlines our dedication to fostering a collaborative, inclusive research community.


\bibliography{anthology,custom}
\bibliographystyle{acl_natbib}

\clearpage
\appendix

\section{Algorithm of \AlgName}

\begin{algorithm}[h]
\caption{\AlgName Workflow}
\label{alg:workflow}
\DontPrintSemicolon
\KwIn{Query set $Q$, training corpus $D$, instruction prompt for keyword assignment $Inst_{\text{key}}$, instruction prompt for supportiveness evaluation $Inst_{\text{eval}}$}
\KwOut{Retrieved Samples $D_{\text{sel}}$}

\BlankLine
\tcc{Stage 1: Semantic Clustering (Offline)}
\BlankLine
$D_{emb} \leftarrow SentenceTransformer(D)$
Leaf Clusters $C=\{c_0, c_1, \ldots, c_{n-1}\} \leftarrow$ Hierarchical clustering on $D_{emb}$ using k-Means (k=10)\;
Semantic Labels $J=\{j_0, j_1, \ldots, j_{n-1}\} \leftarrow \text{LLM}(\{c_0, c_1, \ldots, c_{n-1}\}, Inst_{\text{key}})$\;

\BlankLine
\tcc{Stage 2: Tracing (Online)}
\BlankLine
\For{each query $q \in Q$}{
    $D_{q} \gets \{\}$\;
    $C_{\text{sel}} \leftarrow \text{fuzzymatch}(q, J, C)$ \;
    $Batches \leftarrow$ partition $C_{sel}$ into batches of size $b$
    
    \For{each batch $B \in Batches$}{
        $S_B \leftarrow \text{LLM}(q, B, Inst_{\text{eval}})$\;
        $D_q \gets D_q \cup \{z \mid z \in B, s_i = 1\}$\;
    }
    $D_{\text{sel}} \gets D_{\text{sel}} \cup D_q$
}
\end{algorithm}

\section{Extended Related Work}
\paragraph{Training Data Attribution (TDA).}
TDA aims to trace model predictions back to the training examples that responsible for these predictions. As it shares a similar goal with fact tracing, prior work \citep{akyurek2022towards} proposes to use two popular families of TDA methods as baselines.
Gradient-based attribution, for instance, focuses on quantifying the direct influence a particular training example $z$ has on the loss at a test example $z_\text{query}$, when using a model parameterized by $\theta$. 
A notable technique within this category is \textsc{TracIn} \cite{pruthi2020estimating}. It employs a first-order Taylor approximation to estimate the loss change on $z_\text{query}$ when taking a gradient step on training example $z$ at time $t$. The resulting attribution score is simply a dot product of gradients at a particular step t: 

\vspace{-1em}\begin{equation}
\mathcal{I}_{\mathrm{t}}\left(z, z_{\text {query }}\right)=\nabla_\theta L\left(z_{\text {query }}, \theta_{\mathrm{t}}\right)^{\top} \nabla_\theta L\left(z, \theta_{\mathrm{t}}\right)
\end{equation}\normalsize

Embedding-based attribution employs the model's internal representations to determine the relevance of training examples to a given test prediction \citep{rajani2019explain}. The attribution score is defined as a cosine product of hidden representations:

\vspace{-1em}\begin{equation}
\label{eq: embed}
\mathcal{I}\left(z, z_{\text {query }}\right)= \frac{L M_{\text {inter. }}(z)^{\top} L M_{\text {inter. }}\left(z_{\text {query }}\right)}{\left\|L M_{\text {inter. }}(z)^{\top}\right\|\left\|L M_{\text {inter. }}\left(z_{\text {query }}\right)\right\|}
\end{equation}\normalsize


To retrieve supporting training data for a given query $z_\text{query}$, one need to score every training data and rank them by their influence score. However, TDA methods fail to justify groundeness and often perform worse than simple IR baseline (i.e., BM25) \citep{akyurek2022towards}. Moreover, it could be computationally infeasible for gradient-based TDA scoring all training data in large datasets, and relies on evaluation on carelly selected smallsubset (i.e., around 500) for each query. This limitation motivates us to design a framework that is both more computationally efficient and more effective.

\paragraph{Information Retrieval (IR).} 
IR focuses on retrieving relevant documents in a large collection given specific queries \citep{izacard2021unsupervised}. Though not theoretically justified for fact tracing task, prior work \citep{akyurek2022towards} found it could serve as a possible solution and outperforms principled TDA methods by a large marigin.
There are two branches of IR methods: term-frequency based \citep{mikolov2013efficient, lv2011lower, robertson1995okapi} and neural network based \citep{karpukhin2020dense,xiong2020approximate,izacard2021unsupervised,ni2021large}. 
A classic example of the former one is BM25 \citep{robertson1995okapi, lv2011lower}, which represents the best performing variant of lexical similarity-based IR methods. When using BM25 for fact tracing, one can treat examples as a bag of words, and score each training data base on the token overlap with the given query, inversely weighted with the frequency of such tokens.
On the other hand, neural network-based methods often require labor-intensive annotations on query-document pairs \citep{karpukhin2020dense,xiong2020approximate}. This making them impractical in the fact tracing scenario where such annotations are not available. While some recent works \citep{izacard2021unsupervised,ni2021large} propose to overcome the limitation using zero-shot learning, they usually results in an inferior retrieval quality, even worse than a non-parametric BM25 \citep{thakur2021beir,zhou-etal-2022-hyperlink}. Thus, we follow \citet{akyurek2022towards} and choose BM25 as IR baseline for fact tracing.

Similar to TDA methods, IR methods also focus on \emph{relevance} while neglecting the \emph{supportiveness} of the connection between training data and the query. In this paper, we introduce \AlgName, the first supportiveness-aware approach for fact tracing, offering substantial benefits in scenarios where training data contains conflicting information, such as time-sensitive facts in news reports.



\section{Baseline Implementation Details}\label{sec:baseline-details}
\paragraph{\textsc{TracIn}}
We follow the setup of \citet{akyurek2022towards}, optimize \textsc{TracIn} by recaling gradient with Adafactor accumulators, applying unit-normalization to the gradients and selecting the best performing layer, which is first encoder layer. In practice, it is found that aggregating over multiple checkpoints often does not lead to improved performance but raises the computational burden. Thus, it is preferred to just use a single checkpoint that gives the best results~\citep{just2023lava}. For FTRACE-TREx dataset, we use a MT5 model finetuned on it for 80k steps.

\paragraph{\textsc{Embed}}
We use the same MT5 checkpoint as for \textsc{TracIn} on FTRACE-Trex dataset. For VITAMINC dataset, we finetune a Bert model by randomly masking some tokens of the training data. We observe that best performing layer is the last encoder layer of Bert and use results of this layer as the final results.

\section{\AlgName Implementation Details}

A detailed algorithm of \AlgName is given in \ref{alg:workflow}. When perform hierachical clustering, we employ k-means \citep{na2010research}. The clustering process is applied recursively: clusters with more than $c$ samples will be clustered again until it contains less then $c$ samples. We use $c=100, k=10$ for all experiments. 

For keyword assignment, we set temperature to be 0 and output length to be 256 when calling GPT-4 api.
For the batch process at Stage 2, we use a batch size $b=15$. We set temperture to be 0 and output length to be 1024. Prompts used can be found in Section \ref{sec:prompt}.

\section{More Results}\label{sec:res-more}
\paragraph{BM25's performance before and after paraphrasing queries}
BM25 operates based on token overlap, and retrieves examples with high lexical similarity to the query, regardless of their factual consistency. Its performance can be poor even when there are no such data conflicts. We further conduct experiment on FTRACE-TREx dataset where we paraphrase each query using an open-sourced paraphraser \footnote{\url{https://huggingface.co/humarin/chatgpt_paraphraser_on_T5_base}}. The performance of BM25 before and after paraphrasing is shown in Table \ref{tab:bm25-paraphrase}, where both precision and recall drop by a wide margin.

\begin{table*}[h]
\centering
\caption{BM25's performance before and after paraphrasing queries in FTRACE-TREx dataset. Notably, BM25 exhibits a 21 percentage point drop in precision, while \AlgName maintains consistent performance, achieving a precision of 0.81 and a recall of 0.69.}
\label{tab:bm25-paraphrase}
\scalebox{1}{
\begin{tabular}{c c c c c c c}
\toprule
& \multicolumn{2}{c}{\textbf{Top-1}} & \multicolumn{2}{c}{\textbf{Top-10}} & \multicolumn{2}{c}{\textbf{Top-25}} \\
\cmidrule(lr){2-3} \cmidrule(lr){4-5} \cmidrule(lr){6-7}
& {\textbf{Precision}} & {\textbf{Recall}} & {\textbf{Precision}} & {\textbf{Recall}} & {\textbf{Precision}} & {\textbf{Recall}} \\
\midrule
 \textbf{Before} & 0.83 & 0.06 & 0.66 & 0.36 & 0.49 & 0.52 \\
\textbf{After}  & 0.62 & 0.05 & 0.48 & 0.28 & 0.38 & 0.42 \\
\bottomrule
\end{tabular}}
\end{table*}

\paragraph{\textsc{TracIn}'s performance when using different model checkpoints}
\begin{table*}[h]
\centering
\caption{\textsc{TracIn}'s performance using checkpoints at gradient steps 30k and 80k}
\label{tab:tracin-step}
\scalebox{1}{
\begin{tabular}{c c c c c c c}
\toprule
& \multicolumn{2}{c}{\textbf{Top-1}} & \multicolumn{2}{c}{\textbf{Top-10}} & \multicolumn{2}{c}{\textbf{Top-25}} \\
\cmidrule(lr){2-3} \cmidrule(lr){4-5} \cmidrule(lr){6-7}
& {\textbf{Precision}} & {\textbf{Recall}} & {\textbf{Precision}} & {\textbf{Recall}} & {\textbf{Precision}} & {\textbf{Recall}} \\
\midrule
 \textbf{30k} & 0.10 & 0.003 & 0.02 & 0.01 & 0.01 & 0.01 \\
\textbf{80k}  & 0.19 & 0.01 & 0.05 & 0.02 & 0.03 & 0.03 \\
\bottomrule
\end{tabular}}
\end{table*}

\begin{table*}[h]
\caption{Upperbound performance of \AlgName using different clustering algorithms on FTRACE-TREx. Different embedding models do not bring much effects on the performance upperbound, demonstrating the robustness of \AlgName on the choice of embeddings. *\textit{We list GPU time if it is an open-sourced model deployed on our server and costs if it is accessed through queries to the API.}}
\label{tab: emb}
\centering
\scalebox{1}{
\begin{tabular}{lccc}
\hline
\textbf{Embedding Scheme}        & \textbf{Precision} & \textbf{Recall} & \textbf{Time/Costs*}\\ \hline
Sentence-Transformer        & 0.81               & \textbf{0.65}         & 0.16 min\\
Cohere Clustering         & 0.80               & 0.63       & \$ 0.04   \\
Cohere Classification     & 0.75               & 0.57    & \$ 0.04        \\
Cohere Search-Query       & 0.79               & 0.60  & \$ 0.04    \\
Cohere Search-Document    & \textbf{0.82}               & 0.56   & \$ 0.04  \\
Mistral-Embed        & 0.69               & 0.53   &   \$0.05   \\ \hline
\end{tabular}}
\end{table*}\vspace{-0.5em}

\paragraph{\AlgName's performance with larger corpus size.}
\citet{akyurek2022towards} benchmark tracing methods only on a curated small candidate set with about 500 examples for each query. In contrast, our benchmark has assessed the efficacy and efficiency of our method on a significantly larger corpus, containing 10k instances.
Moving forward, we aim to tackle a more challenging scenario: we use the same query set of VITAMINC, but augment the attribution set with additional non-relevant examples until the total reaches 100k. This setting is designed to evaluate our method's robustness in scenarios that better resemble real-world applications. As shown in Table \ref{tab:corpus}, both methods exhibit a slight decline in performance, yet \AlgName consistently outperforms BM25 by a significant margin. 

BM25's performance drop is ascribed to the inclusion of new examples that exhibit high lexical overlap with the queries, thereby impairing BM25's effectiveness. For \AlgName, the performance drop is primarily observed in the initial stage, where some clusters are not successfully retrieved. This results from the clustering logic changes after more diverse sample are included. Specifically, if a cluster contains only a few groundtruth examples, these may be overlooked during the semantic labeling process, leading to false negative retrieval. Despite this minor reduction in performance, \AlgName consistently outperforms BM25 by a significant margin.

\begin{table*}[h]
\centering
\caption{Performance of BM25 and \AlgName when dealing with different corpus size. Both of the methods encounter a slightly performance drop, while \AlgName still $1.66\times$ better than BM25.}
\scalebox{1}{
\begin{tabular}{ccccccc}
\hline
\textbf{}      & \multicolumn{3}{c}{\textbf{VITAMINC-10k}}          & \multicolumn{3}{c}{\textbf{VITAMINC-100k}}         \\ \cmidrule(lr){2-4} \cmidrule(lr){5-7}
\textbf{}      & \textbf{F1} & \textbf{Precision} & \textbf{Recall} & \textbf{F1} & \textbf{Precision} & \textbf{Recall} \\ \hline
\textbf{BM25}  & 0.55        & 0.59               & 0.53            &  0.53           & 0.56                   & 0.50                \\
\textbf{Ours}  & 0.91        & 0.88               & 0.98            &  0.88           &   0.85                 & 0.92               \\
\textbf{Ours\textsuperscript{*}} & 1.00        & 1.00               & 1.00            &  0.95           & 0.95                   & 0.95                \\ \hline
\end{tabular}}
\label{tab:corpus}
\end{table*}\vspace{-0.5em}

\paragraph{Failure example of \AlgName.}
Example below shows that GPT-4 struggles with complex logical reasoning involving dates; for instance, it incorrectly equates the information from different dates, focusing merely on numerical comparisons.

\begin{tcolorbox}[rounded corners, colback=lightgray!20, colframe=black, boxrule=0.5pt, 
    top=2mm, bottom=2mm, left=2mm, right=2mm]
\small
\textbf{Query}: As of 14 March , there have been more than 20 confirmed COVID-19 cases in Kerala .\\[2ex]
\textbf{False positive examples}:\\
1: 2020 coronavirus pandemic in Kerala: As of 28th March 2020 , there are 182 confirmed cases of the virus and more than 1,15,000 people are under surveillance in the state .\\
2: 2020 coronavirus pandemic in Kerala: As of 29th March 2020 , there are 202 confirmed cases of the virus and more than 1,41,000 people are under surveillance in the state .\\[2ex]
\textbf{GPT-4 analysis:}\\
Text 1: States that "As of 28th March 2020, there are 182 confirmed cases of the virus in Kerala." This text supports the claim because if there were 182 cases by the 28th of March, there must have been more than 20 cases by the 14th of March.\\
Text 2: States that "As of 29th March 2020, there are 202 confirmed cases of the virus in Kerala." This text also supports the claim because if there were 202 cases by the 29th of March, there must have been more than 20 cases by the 14th of March.
\end{tcolorbox}

\section{Prompting Templates}\label{sec:prompt}

\begin{figure*}[ht]
\caption{Prompt template for keyword assignment.}
\begin{tcolorbox}[sharp corners, colback=blue!10, colframe=black, boxrule=0.5pt, 
    top=2mm, bottom=2mm, left=2mm, right=2mm]
\textbf{Prompt:} \\[2ex]
Analyze the following group of sentences and identify 5 to 10 of phrases that capture the main topics, focusing on the key entities.

Group of sentences: [cluster texts]

Output the keywords in the following format:
\#keywords: your keywords here.
[note: Extract 5 to 10 keywords only. The keywords should capture the main topic and seperated by comma.]

\end{tcolorbox}
\end{figure*}

\begin{figure*}[ht]
\caption{Prompt template for supportiveness evaluation on FTRACE-TREx dataset.}
\begin{tcolorbox}[sharp corners, colback=blue!10, colframe=black, boxrule=0.5pt, 
    top=2mm, bottom=2mm, left=2mm, right=2mm]
\textbf{Prompt:} \\[2ex]
I will give you a claim and multiple texts. Carefully evaluate each text, check if the text supports the claim. \\
    \\
For example, \\
\\
Fact: Member of the Scottish Parliament is a legal term in Scotland.\\ 
\\
Group of Texts: \\
Text 1: Dennis Robertson is a Scottish politician, and has been an Member of the Scottish Parliament (MSP) for Aberdeenshire West since 2011, after defeating the Liberal Democrat incumbent, Mike Rumbles, by a majority of 4,112 votes. \\
Text 2: The West Lothian question, also known as the English question, refers to whether MPs from Northern Ireland, Scotland and Wales, sitting in the House of Commons of the United Kingdom, should be able to vote on matters that affect only England, while MPs from England are unable to vote on matters that have been devolved to the Northern Ireland Assembly, the Scottish Parliament and the Welsh Assembly.
\\
\#analysis: A "legal term" refers to a term or expression that is associated with or used in the formal context of a particular country. Text1 mentions that "Member of the Scottish Parliament (MSP)" is in Scotland; because we can infer that member of the scottish parliament is used in the formal political context of Scotland, it implicitly establishes that Member of the Scottish Parliament is a legal term in Scotland. Text2 mentions the Scottish Parliament but does not state that "Member of the Scottish Parliament" is a legal term used within the context of the Scotland. \\
\#scores: 1, 0\\
\\
Fact: [query]\\
Group of Texts: [indexed candidate training data]\\
\\
Now evaluate each text carefully in the group and output in the following format:\\
\\
\#analysis: your analysis here.\\
\#scores: your scores here. (score each text 0 or 1 according to the analysis.)\\
\end{tcolorbox}
\end{figure*}

\begin{figure*}[ht]
\caption{Prompt template for supportiveness evaluation on VITAMINC dataset.}
\begin{tcolorbox}[sharp corners, colback=blue!10, colframe=black, boxrule=0.5pt, 
    top=2mm, bottom=2mm, left=2mm, right=2mm]
\textbf{Prompt:} \\[2ex]
I will give you a claim and multiple texts. Carefully evaluate each text, check if the text supports the claim.\\
    \\
For example, \\
\\
Claim: Black Mass earned less than \$ 43.6 million in North America as of 22 April 2020.\\

Group of Texts: \\
Text 1: Black Mass has grossed less than \$ 42.6 million in North America as of 22 April 2020.\\
Text 2: Black Mass has grossed less than \$ 42.6 million in North America as of 12 April 2020.\\
Text 3: Black Mass has grossed more than \$ 42.6 million in North America as of 22 April 2020.\\
Text 4: Black Mass has grossed more than \$ 43.6 million in North America as of 22 April 2020.\\

\#analysis: \\
Text 1: States that "Black Mass has grossed less than \$42.6 million in North America as of 22 April 2020." This text supports the claim because if it grossed less than \$42.6 million, it also grossed less than \$43.6 million. \\
Text 2: States that "Black Mass has grossed less than \$42.6 million in North America as of 12 April 2020." This text does not directly suppport or refute the claim because it provides information as of 12 April. Without specific information on the movie's earnings trends or events that might have affected its box office performance between 12 April and 22 April 2020, it is impossible to determine whether the gross was less than \$43.6 million as of 22 April.\\
Text 3: States that "Black Mass has grossed more than \$42.6 million in North America as of 22 April 2020." This text does not directly support or refute the claim because it does not provide enough information to determine whether the gross was less than \$43.6 million. It only indicates that the gross was more than \$42.6 million, which could still be less than \$43.6 million. \\
Text 4: States that "Black Mass has grossed more than \$43.6 million in North America as of 22 April 2020." This text does not support the claim because it directly contradicts it, indicating that the gross was more than \$43.6 million.\\

\#scores: 1, 0, 0\\
\\
Fact: [query]\\
Group of Texts: [indexed candidate training data]\\
\\
Now evaluate each text carefully in the group and output in the following format:\\
\\
\#analysis: your analysis here.\\
\#scores: your scores here. (score each text 0 or 1 according to the analysis.)\\
\end{tcolorbox}
\end{figure*}

\end{document}